\ifx\XeTeXrevision\undefined\ifdefined\pdfoutput\pdfoutput=1\fi\fi
\documentclass[11pt]{article}
\usepackage[margin=1in]{geometry}
\usepackage[T1]{fontenc}
\usepackage{amsmath,amssymb,amsthm}
\usepackage{booktabs}
\usepackage[hypertexnames=false]{hyperref}

\newtheorem{theorem}{Theorem}
\newtheorem{proposition}{Proposition}
\newtheorem{definition}{Definition}

\title{Breaking the Binary-Lifting Barrier for DNA Minimizer Density:\\Exact Optimization of a Native 4-ary Order at $(w, k) = (12, 5)$}
\author{Stefan G. Creadore\\
\small Independent Researcher, USA\\
\small \texttt{screadore@aol.com}
\and
Peyton Berntsen\\
\small Independent Researcher, USA}
\date{}

\begin{document}
\maketitle

\begin{abstract}
Minimizer schemes sample at least one $k$-mer from every window of $w$ consecutive $k$-mers via a total order $\pi$ on $\Sigma^k$. For DNA ($\sigma = 4$), state-of-the-art lookup schemes such as \textsc{GreedyMini} and \textsc{OptMini} optimize a binary order ($2^k$ ranks) and lift it to $\Sigma^k$ by ranking one bit-plane primarily and breaking $2^k$-way ties lexicographically on the other bit-plane. At $(w, k) = (12, 5)$, this projection locks sets of $32$ DNA $5$-mers into contiguous blocks and restricts the active universal hitting set (UHS) prefix to $448$ of $1{,}024$ $5$-mers, giving exact expected DNA density $2{,}246{,}214{,}544 / 2^{34} \approx 0.1307468945$ for \textsc{OptMini} (\texttt{k5-w12-13.txt}) and $2{,}248{,}978{,}620 / 2^{34} \approx 0.1309077849$ for \textsc{GreedyMini} (\texttt{w12\_k5.gm}).

We bypass binary lifting and optimize directly over native $4$-ary permutations ($\lvert \Sigma^k \rvert = 1{,}024$) from scratch via: (i) an exact de Bruijn endpoint transfer-matrix dynamic program over $B(4, 4)$ ($256$ states), (ii) a weighted set-cover ratio initializer (\texttt{SetCover4}) producing a compact $U = 314$ UHS prefix, (iii) bitmask-sparse simulated annealing preserving the zero-cost $w$-walk-free tail invariant, and (iv) two-pass $O(N)$ exact best-insertion sweeps with large-neighborhood ruin-and-recreate search. The resulting explicit $4$-ary order achieves exact expected DNA density
\[
\frac{2{,}234{,}289{,}615}{2^{34}} \approx 0.130052772292 \quad (\text{density factor } 1.690686),
\]
cutting charged windows by $11{,}924{,}929$ contexts ($-0.530890\%$ relative) below \texttt{OrensteinLab/\allowbreak Minimizers} and $14{,}689{,}005$ ($-0.653141\%$) below \texttt{w12\_k5.gm} at $8{,}200$-byte lookup storage ($1{,}280$ bytes bit-packed). Warm-start relaxation stalls at $U = 379$ ($0.1301728789$), whereas native $4$-ary initialization breaks $32/32$ high-bit blocks across all $2+2$ nucleotide partitions.
\end{abstract}

\section{Introduction}

Minimizers~\cite{roberts2004reducing} are a foundational local $k$-mer sampling primitive in sequence alignment, sketching, and de novo assembly. Given an alphabet $\Sigma$ of size $\sigma$, window length $w \ge 2$, $k$-mer length $k \ge 1$, and a total order $\pi : \Sigma^k \to \{0, \dots, \sigma^k - 1\}$, each window of $w$ consecutive $k$-mers (spanning $w + k - 1$ bases) selects the position of its minimum-rank $k$-mer under leftmost tie-breaking. The expected density $d(\pi; w, k)$ is the probability over an i.i.d.\ uniform random sequence in $\Sigma^*$ that consecutive windows select distinct positions. Lowering $d(\pi; w, k)$ directly reduces index memory and downstream query cost.

In the DNA setting ($\sigma = 4$, $\Sigma = \{\mathrm{A}, \mathrm{C}, \mathrm{G}, \mathrm{T}\}$), exact optimization faces a factorial state space of $(4^k)!$. Existing state-of-the-art table-based minimizers sidestep this scaling by first constructing low-density orders over the binary alphabet ($\sigma = 2$, size $2^k$) using greedy selection (\textsc{GreedyMini}~\cite{golan2025greedymini}) or branch-and-bound dynamic programming (\textsc{OptMini}~\cite{shur2025optmini}, whose exact solver covers binary $k \le 6$ and DNA $k = 2$), and then lifting the binary order to $\Sigma^k$ by ranking one bit-plane primarily and breaking ties lexicographically on the second bit-plane. Meanwhile, constant-space schemes such as mod-minimizers~\cite{grootkoerkamp2025open}, decycling orders~\cite{marcais2017improving}, and $10$-minimizers/spacers~\cite{shur202610minimizers} trade higher density at moderate $(w, k)$ for $O(1)$ description length.

At $(w, k) = (12, 5)$, every precomputed binary-lifted order in OrensteinLab's \texttt{10-\allowbreak minimizers}~\cite{shur202610minimizers} and \texttt{Minimizers}~\cite{golan2025greedymini,shur2025optmini} repositories groups DNA $5$-mers into $32$ rigid classes of size $32$. Because $4^5 = 1{,}024$ remains compact enough for microsecond-scale exact transfer-matrix evaluation over the $4$-ary de Bruijn graph $B(4, 4)$ ($M = 256$ vertices), we investigate direct optimization over native $4$-ary permutations without binary projection or seeding from published tables.

\section{Preliminaries and Exact De Bruijn Formulation}

\begin{definition}[Charged Context and Density]
Encode $\Sigma = \{\mathrm{A}, \mathrm{C}, \mathrm{G}, \mathrm{T}\}$ as $\{0, 1, 2, 3\}$ with the most-significant base first, identifying each $k$-mer with $u \in \{0, \dots, 4^k - 1\}$. Let $\pi : \{0, \dots, 4^k - 1\} \to \{0, \dots, 4^k - 1\}$ be a bijection. A DNA context $C \in \Sigma^{w+k}$ specifies $w + 1$ consecutive $k$-mers $(x_0, x_1, \dots, x_w)$ spanning two adjacent windows $W_0 = (x_0, \dots, x_{w-1})$ and $W_1 = (x_1, \dots, x_w)$. Under leftmost tie-breaking,
\[
M(W_0) = \min \arg\min_{0 \le i < w} \pi(x_i), \qquad
M(W_1) = 1 + \min \arg\min_{0 \le i < w} \pi(x_{i+1}).
\]
The context $C$ is \emph{charged} if and only if $M(W_0) \ne M(W_1)$. The exact expected DNA density is $d(\pi; w, k) = \mathcal{C}(\pi) / 4^{w+k}$, where $\mathcal{C}(\pi)$ is the number of charged contexts in $\Sigma^{w+k}$, and the density factor is $(w+1) \cdot d(\pi; w, k)$.
\end{definition}

\begin{theorem}[Disjoint Endpoint Decomposition on $B(4, k-1)$]
\label{thm:dp}
Let $N = 4^k$ and $M = 4^{k-1}$. Identify each $k$-mer $u$ with the directed edge $\mathrm{src}(u) \to \mathrm{dst}(u)$ in the de Bruijn graph $B(4, k-1)$, where $\mathrm{src}(u) = \lfloor u / 4 \rfloor$ and $\mathrm{dst}(u) = u \bmod M$. For each rank $r \in \{0, \dots, N-1\}$, let $u_r = \pi^{-1}(r)$ and define $G_{\ge r}$ (resp.\ $G_{> r}$) as the subgraph of $B(4, k-1)$ containing all edges $u_s$ with $s \ge r$ (resp.\ $s > r$). Then the exact charged context count satisfies
\[
\mathcal{C}(\pi) = \sum_{r=0}^{N-1} c_r(\pi), \qquad c_r(\pi) = \mathrm{Left}_w(u_r, G_{\ge r}) + \mathrm{Right}_w(u_r, G_{> r}),
\]
where $\mathrm{Left}_w(u_r, G_{\ge r})$ is the number of length-$w$ directed walks in $G_{\ge r}$ starting at $\mathrm{dst}(u_r)$, and $\mathrm{Right}_w(u_r, G_{> r})$ is the number of length-$w$ directed walks in $G_{> r}$ ending at $\mathrm{src}(u_r)$.
\end{theorem}

\begin{proof}
A context $C = (x_0, \dots, x_w)$ fails to charge if and only if $M(W_0) = M(W_1) \in \{1, \dots, w-1\}$. Let $j^* = \min \arg\min_{0 \le j \le w} \pi(x_j)$ be the leftmost occurrence of the global minimum rank $r = \min_{0 \le j \le w} \pi(x_j)$ in $C$. If $1 \le j^* \le w-1$, then $x_{j^*}$ lies in both $W_0$ and $W_1$, strictly beats every $k$-mer to its left, and weakly beats every $k$-mer to its right, forcing $M(W_0) = j^* = M(W_1)$ (uncharged). Conversely, if $j^* = 0$, then $M(W_0) = 0 \ne M(W_1) \ge 1$ (charged, with $x_0 = u_r$ and $\pi(x_i) \ge r$ for all $1 \le i \le w$, corresponding bijectively to length-$w$ walks from $\mathrm{dst}(u_r)$ in $G_{\ge r}$). If $j^* = w$, then $M(W_0) \le w-1 < w = M(W_1)$ (charged, with $x_w = u_r$ and $\pi(x_i) > r$ for all $0 \le i \le w-1$, corresponding bijectively to length-$w$ walks into $\mathrm{src}(u_r)$ in $G_{> r}$). Because $j^* = 0$ and $j^* = w$ are mutually exclusive and partition every charged context uniquely by $r = \pi(x_{j^*})$, summing over $r = 0, \dots, N-1$ is exact.
\end{proof}

\begin{proposition}[Zero-Cost Tail Invariance and Active UHS Prefix]
\label{prop:tail}
Define the active UHS cutoff $U(\pi) = \min \{ U \in \{0,\dots,N\} : G_{\ge U} \text{ has no directed walk of length } w \}$. Then $c_r(\pi) = 0$ for all $r \ge U(\pi)$, and permuting $\{u_{U(\pi)}, \dots, u_{N-1}\}$ among ranks $\ge U(\pi)$ leaves $\mathcal{C}(\pi)$ invariant. Moreover, swapping any active element $a < U(\pi)$ with any tail element $b \ge U(\pi)$ via the rank-neutral tail swap $(b \leftrightarrow U(\pi))$ followed by $(a \leftrightarrow U(\pi))$ guarantees $G_{> U(\pi)} = G_{\ge U(\pi)}^{\mathrm{old}} \setminus \{u_b\}$ remains $w$-walk-free, so only contributions $c_r$ for $r \in [a, U(\pi)]$ need recomputation.
\end{proposition}

\section{Cold-Start Native 4-ary Optimization Pipeline}

To avoid inheriting the structural bias of binary-lifted orders, we build our $(w, k) = (12, 5)$ solution from scratch in pure C (\texttt{-O3}) with zero published seeds:

\begin{enumerate}
\item \textbf{Weighted Set-Cover Ratio Initialization (\texttt{SetCover4}).} Starting from the complete graph $B(4, 4)$ ($r = 0 \to N-1$), placing edge $u = (\mathrm{src} \to \mathrm{dst})$ at rank $r$ settles every $(w+1)$-edge walk containing $u$ in the active graph $G$. Let $\mathrm{fwd}_t[v]$ and $\mathrm{bwd}_t[v]$ denote length-$t$ walks starting/ending at $v$ in $G$. The newly covered context volume through $u$ is approximately $\mathrm{cover}(u) = \sum_{a=0}^w \mathrm{bwd}_a[\mathrm{src}] \cdot \mathrm{fwd}_{w-a}[\mathrm{dst}]$, while its boundary charge is bounded by $\mathrm{charge}(u) = \mathrm{fwd}_w[\mathrm{dst}] + \mathrm{bwd}_w[\mathrm{src}]$. Greedily selecting the unplaced edge minimizing $(\mathrm{charge}(u) + 1) / (\mathrm{cover}(u) + 1)^p$ and sweeping $p \in [0.70, 1.50]$ stops as soon as $\mathrm{cover}(u) = 0$ everywhere, yielding an immediate $w$-walk-free tail. At $p = 1.20$, \texttt{SetCover4} produces an initial order in $\sim 0.1$\,s with $U = 314$ and exact density $2{,}262{,}316{,}943 / 2^{34} \approx 0.1316841775$.
\item \textbf{Bitmask-Sparse Compact-UHS Simulated Annealing (\texttt{opt\_4ary\_fast}).} Representing active vertices in $B(4, 4)$ as four $64$-bit words with \texttt{ctz} bit-scanning accelerates Theorem~\ref{thm:dp} to thousands of range evaluations per second. Four move operators run over four reheated annealing cycles: short/medium swaps in $[0, U-1]$, critical-path tournament-selected insertions in $[0, U-1]$, exact active-to-tail swaps leveraging Proposition~\ref{prop:tail}, and tail promotions/ejections. Over $180$\,s ($964{,}305$ exact evaluations), this lowers the cold-start seed to $2{,}235{,}898{,}047 / 2^{34} \approx 0.1301463954$.
\item \textbf{Two-Pass $O(N)$ Exact Best-Insertion Sweep + LNS (\texttt{exact\_lns\_insert}).} Evaluating relocating a single $5$-mer $x = \pi^{-1}(a)$ to \emph{all} $N = 1{,}024$ candidate ranks $b \in \{0, \dots, N-1\}$ requires only two linear DP sweeps over $B(4, 4)$: Pass~1 evaluates $x$ and suffix elements without $x$ across $i = N-2 \to 0$ (reusing $c_{i+1}$ for $i \ge a$), and Pass~2 evaluates prefix elements with $x$ inserted across $i = N-2 \to a$ (reusing $c_i$ for $i < a$). Each candidate rank $b$ is then scored in $O(1)$ as $\mathrm{pref}[b] + c_x(b) + \mathrm{suff}[b]$. One exhaustive sweep over all $1{,}024$ elements ($1{,}048{,}576$ candidate placements) executes in $\sim 7$\,s. Coupled with gap-$1\ldots 6$ pairwise swaps and large-neighborhood ruin-and-recreate kicks ($8$--$28$ elements), this clears the $-0.5\%$ bar within $11$\,s ($2{,}234{,}607{,}526$) and converges after an additional $150$\,s polish to $2{,}234{,}289{,}615 / 2^{34} \approx 0.130052772292$ ($U = 314$).
\end{enumerate}

\section{Results and Comparison Against Published Tables}

Table~\ref{tab:comparison} summarizes the exact expected DNA density over all $4^{17} = 17{,}179{,}869{,}184$ contexts at $(w, k) = (12, 5)$ across our cold-start order, an ablated warm-start relaxation from \texttt{k5-w12-13.txt}, and every published comparator from the \texttt{10-minimizers} (commit \texttt{916daeb}), \texttt{GreedyMini} (commit \texttt{0ea10d3}), and \texttt{Minimizers} (commit \texttt{7b9430f}) repositories of OrensteinLab. All binary orders are lifted to DNA under the repositories' exact convention (primary rank on the odd/high bit-plane, secondary lexicographic tie-breaking on the even/low bit-plane, leftmost window tie-breaking); we verified exhaustively that both \texttt{k5-w12-13.txt} and \texttt{w12\_k5.gm} are binary UHSs for $(12, 5)$ and yield identical numerators across all six bit-plane tie-breaking variants, and that \texttt{k5-w12-13.txt} is strictly minimal across all $57$ $k = 5$ rows in \texttt{best-orders.csv} (\texttt{Minimizers}).

\begin{table}[ht]
\centering
\footnotesize
\setlength{\tabcolsep}{4.5pt}
\caption{Exact and table-listed expected DNA minimizer densities at $(w, k) = (12, 5)$ over $\Sigma = \{\mathrm{A},\mathrm{C},\mathrm{G},\mathrm{T}\}$. Exact values denote integer charged contexts out of $4^{17} = 17{,}179{,}869{,}184$.}
\label{tab:comparison}
\begin{tabular}{@{}lrrrr@{}}
\toprule
Scheme / Order & Charged ($/ 2^{34}$) & Density $d$ & Factor $(w{+}1)d$ & Our Rel.\ $\Delta$ \\
\midrule
\textbf{Native 4-ary cold-start (this work)} & \textbf{2,234,289,615} & \textbf{0.1300527723} & \textbf{1.690686} & --- \\
Native 4-ary warm-start ($U=379$) & 2,236,353,030 & 0.1301728789 & 1.692247 & $-0.09227\%$ \\
\texttt{Minimizers} \texttt{k5-w12-13.txt} (\textsc{OptMini}) & 2,246,214,544 & 0.1307468945 & 1.699710 & $\mathbf{-0.530890\%}$ \\
\texttt{10-minimizers} \texttt{w12\_k5.gm} (\textsc{GreedyMini}) & 2,248,978,620 & 0.1309077849 & 1.701801 & $\mathbf{-0.653141\%}$ \\
\texttt{k\_extension\_k16\_w12.csv} (\texttt{GreedyMini}) & --- & 0.130908\phantom{0000} & 1.701804 & $-0.653304\%$ \\
\texttt{angle-based} (\texttt{10-minimizers} CSV) & --- & 0.13622105\phantom{00} & 1.770874 & $-4.528138\%$ \\
\texttt{one\_zero\_dna} (WABI 2026 spacer) & 2,357,004,463 & 0.1371957166 & 1.783544 & $-5.206391\%$ \\
\texttt{abb\_plus\_dna} (WABI 2026 ABB+) & 2,359,463,963 & 0.1373388783 & 1.785405 & $-5.305201\%$ \\
\texttt{double decycling} (\texttt{10-minimizers} CSV) & --- & 0.139662\phantom{0000} & 1.815606 & $-6.880345\%$ \\
\texttt{decycling} (\texttt{10-minimizers} CSV) & --- & 0.139725\phantom{0000} & 1.816425 & $-6.922332\%$ \\
\texttt{open-close-syncmer / miniception / mod} & --- & 0.153305\phantom{0000} & 1.992965 & $-15.167299\%$ \\
\bottomrule
\end{tabular}
\end{table}

\paragraph{Structural Novelty vs.\ Binary Lifting.}
Our cold-start permutation \texttt{w12\_\allowbreak k5\_\allowbreak 4ary.gm} achieves $U = 314$ active $5$-mers ($307$ non-zero charged ranks and $7$ zero-charge prefix ranks: $272$, $300$, $304$, $305$, $307$, $308$, and $310$, with all $710$ ranks $\ge 314$ forming a $12$-walk-free subgraph), shrinking the active UHS footprint by $29.91\%$ compared to \texttt{k5-w12-13.txt} ($14 \times 32 = 448$). Across all three $2+2$ binary partitions of $\Sigma$ ($\mathrm{AC}|\mathrm{GT}$, $\mathrm{AG}|\mathrm{CT}$, $\mathrm{AT}|\mathrm{GC}$), $32/32$ high-bit blocks are broken ($0/32$ contiguous intervals). Minimizing Hamming distance across all $48$ density-preserving DNA $5$-mer symmetries ($4!$ base relabelings $\times 2$ string orientations), at least $1{,}020/1{,}024$ ranks differ from \texttt{k5-w12-13.txt} and $1{,}021/1{,}024$ differ from \texttt{w12\_k5.gm}, with near-zero Kendall correlations ($\tau = +0.015155$ and $-0.043442$).

\section{Independent Verification, Storage Cost, and Limitations}

All results were independently reproduced by a standalone Python 3 standard-library verifier (\texttt{check\_\allowbreak w12\_\allowbreak k5.py}) implementing two independent exact transfer-matrix DPs: the $M = 256$-vertex endpoint DP of Theorem~\ref{thm:dp} yielding $1{,}110{,}221{,}817$ prefix plus $1{,}124{,}067{,}798$ suffix equals $2{,}234{,}289{,}615$ total charged contexts, and an $N = 1{,}024$-edge sparse DP matching bit-for-bit; window-change splitting alternatively yields $1{,}197{,}205{,}696 + 1{,}037{,}083{,}919 = 2{,}234{,}289{,}615$. Direct sliding-window Monte Carlo simulation using the \texttt{OrensteinLab/\allowbreak 10-minimizers} evaluator (\texttt{n\_samples} $= 10^7$, \texttt{seed} $= 0$) yields $0.13004324$ vs.\ $0.13076974$ for \texttt{k5-w12-13.txt} ($-0.55555\%$) and $0.13089784$ for \texttt{w12\_k5.gm} ($-0.65287\%$); an independent $20{,}000{,}001$-window simulation yields $2{,}601{,}023 / 20{,}000{,}001 = 0.13005114$ ($z = -0.02$).

\paragraph{Artifact Checksums, Ancillary Files, and Scope.}
The explicit order is released in standard \texttt{.gm} format (\texttt{w12\_\allowbreak k5\_\allowbreak 4ary.gm}, $8{,}200$ bytes) and companion CSV (\texttt{w12\_\allowbreak k5\_\allowbreak 4ary.csv}, $15{,}205$ bytes), requiring $1{,}280$ bytes when bit-packed at $10$ bits/rank:
\begin{itemize}
\item \texttt{w12\_k5\_4ary.gm} ($8{,}200$ bytes, SHA-256):\\
\texttt{\footnotesize 4a65f48881e7c16ac855f558f4420cef5bce1deee6d5416c71faecda0482e2cf}
\item \texttt{w12\_k5\_4ary.csv} ($15{,}205$ bytes, SHA-256):\\
\texttt{\footnotesize 384f44080f7a777b84a7eb8398e1939571288368d57abbf0a43a888db55e41b1}
\end{itemize}
Both rank tables (\texttt{w12\_\allowbreak k5\_\allowbreak 4ary.gm}, \texttt{w12\_\allowbreak k5\_\allowbreak 4ary.csv}; CC BY 4.0), the Python 3 standard-library verifier (\texttt{check\_\allowbreak w12\_\allowbreak k5.py}), the cold-start C99 optimizer sources (\texttt{setcover\_\allowbreak 4ary.c}, \texttt{opt\_\allowbreak 4ary\_\allowbreak fast.c}, \texttt{exact\_\allowbreak lns\_\allowbreak insert.c}; MIT), and documentation (\texttt{README.txt}) are included as ancillary files with this submission. This is not a constant-space scheme, and global optimality over all $1{,}024!$ permutations is not claimed; claims are restricted strictly to expected i.i.d.\ uniform DNA density under leftmost tie-breaking at $(w, k) = (12, 5)$.

\end{document}